\documentclass{article}
\usepackage{ijcai21}

\usepackage{times}
\usepackage{soul}
\usepackage{url}
\usepackage[hidelinks]{hyperref}
\usepackage[utf8]{inputenc}
\usepackage[small]{caption}
\usepackage{graphicx}
\usepackage{amsmath}
\usepackage{amsthm}
\usepackage{booktabs}
\usepackage{algorithm}
\usepackage{algorithmic}
\usepackage{subcaption}
\usepackage{comment}
\usepackage{amsmath,amssymb} % define this before the line numbering.
\usepackage{color}
\usepackage{multirow}
\usepackage{algorithm}
\usepackage{algorithmic}
\usepackage{wrapfig}
\usepackage{color}
\usepackage[switch]{lineno} 
\title{Planning with Learned Dynamic Model for Unsupervised Point Cloud Registration}
\author{
Haobo Jiang 
\and
Jin Xie\footnote{Corresponding Author}
\and
Jianjun Qian\And
Jian Yang$^*$
\affiliations
\normalsize{
PCA Lab, Key Lab of Intelligent Perception and Systems for High-Dimensional Information of Ministry of Education\\
Jiangsu Key Lab of Image and Video Understanding for Social Security\\
School of Computer Science and Engineering, Nanjing University of Science and Technology, China} \\
\{jiang.hao.bo, csjxie, csjqian, csjyang\}@njust.edu.cn
\emails
}

\begin{document}
\maketitle
\begin{abstract}
Point cloud registration is a fundamental problem in 3D computer vision. 
In this paper, we cast point cloud registration into a planning problem in reinforcement learning, which can seek the transformation between the source and target point clouds through trial and error. 
By modeling the point cloud registration process as a Markov decision process (MDP), we develop a latent dynamic model of point clouds, consisting of a transformation network and evaluation network. 
The transformation network aims to predict the new transformed feature of the point cloud after performing a rigid transformation (i.e., action) on it while the evaluation network aims to predict the alignment precision between the transformed source point cloud and target point cloud as the reward signal.  
Once the dynamic model of the point cloud is trained, we employ the cross-entropy method (CEM) to iteratively update the planning policy by maximizing the rewards in the point cloud registration process. Thus, the optimal policy, i.e., the transformation between the source and target point clouds, can be obtained via gradually narrowing the search space of the transformation. 
Experimental results on ModelNet40 and 7Scene benchmark datasets demonstrate that our method can yield good registration performance in an unsupervised manner.
\end{abstract}

\section{Introduction}
Point cloud registration plays an important role in a variety of 3D computer vision applications such as object pose estimation~\cite{wong2017segicp}, Lidar SLAM~\cite{deschaud2018imls}, 3D scene reconstruction~\cite{schonberger2016structure}.   
Given source and target point clouds, it aims to search for an optimal rigid transformation consisting of a rotation matrix and a translation vector so that the transformed source point cloud can align the target point cloud optimally. However, due to the large pose changes and occlusions of objects in real scenarios, point cloud registration is still a challenging problem.

In past decades, much research efforts have been dedicated to the point cloud registration task. 
The traditional registration methods such as iterative closest point algorithm (ICP)~\cite{besl1992method} and its variants \cite{1047997,dong2014lietricp} search for the optimal rigid transformation via alternately estimating the correspondences and executing the least squares optimization. 
Nevertheless, due to the non-convexity of the optimization function, the performance of ICP mainly depends on a good initialized transformation and is prone to the local minima with an improper initialization.
To alleviate it, Go-ICP and other variants \cite{yang2013go,fitzgibbon2003robust} employ the branch and bound strategy or other heuristic methods to globally search for the transformation. 
However, these methods usually need much more inference time than the original ICP and just can achieve limited performance improvement. 
In recent years, benefiting from the discriminative representation ability of deep learning, deep point cloud registration methods \cite{gojcic2019perfect,khoury2017learning} can achieve the impressive registration precision. However, their successes mainly rely on large amounts of ground truth corresponding points, which may increase the cost and limit its applications.

\begin{figure}
	\centering
	\begin{subfigure}[b]{1.0\linewidth}
		\includegraphics[width=\linewidth]{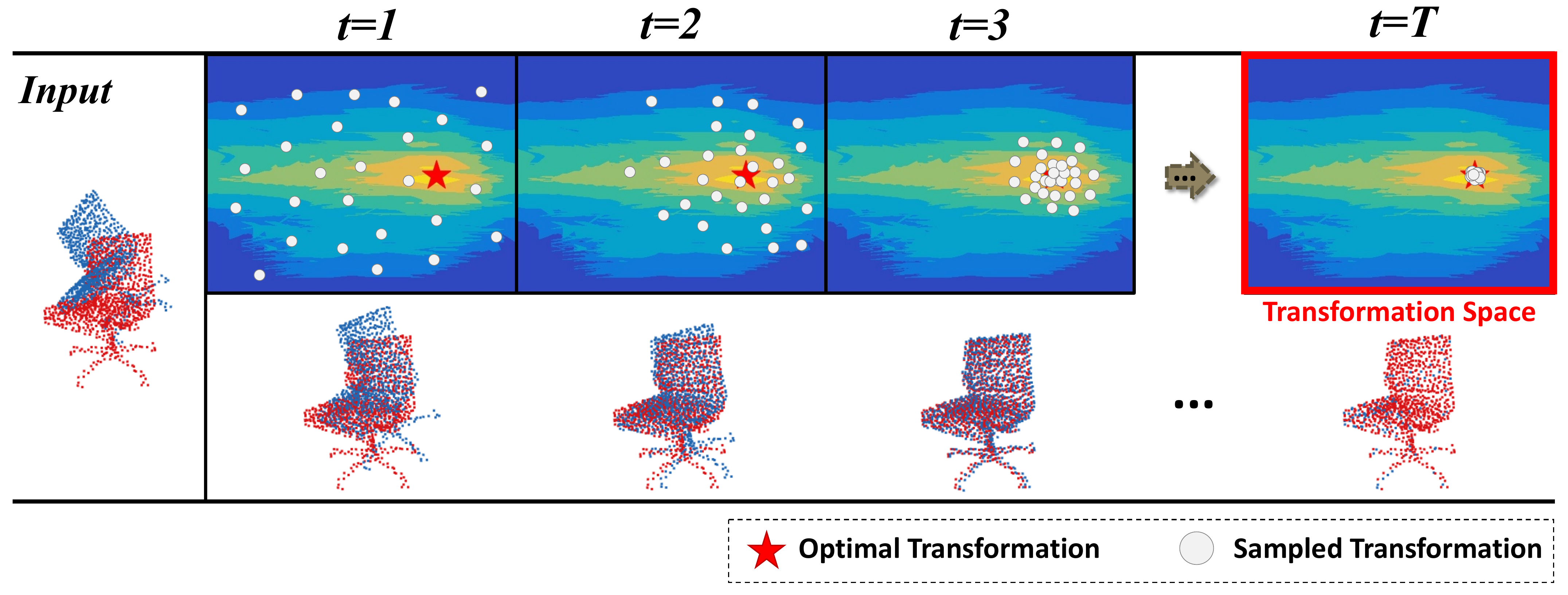}
	\end{subfigure}
	\caption{Planning process of our method for point cloud registration. Our method iteratively narrows the search space  via trial and error (grey points), and iteratively converges to the optimal transformation. The region with the light color denotes the transformations with the high registration precision.}
	\label{insight}
\end{figure}

Inspired by model-based reinforcement learning \cite{hafner2019dream}, in this paper, we propose a novel unsupervised point cloud registration method by converting the point cloud registration problem into a planning problem. With a learned latent dynamic model, it can efficiently seek the rigid transformation between the source and target point clouds by iteratively narrowing the search space of the transformation via trial and error as shown in Fig.~\ref{insight}.
%Based on the Markov decision process, we search for the optimal rigid transformation through planning which 
%Based on the Markov decision process, we first construct a latent dynamic model of point clouds, which can capture the geometric differences between a pair of point clouds in the feature space by performing a rigid transformation on the point clouds. 
Specifically, the constructed dynamic model contains a transformation network and evaluation network. 
The transformation network captures the feature changes when a point cloud is transformed, and is used to predict the new transformed feature of the point cloud after performing a rigid transformation on it.
The evaluation network captures the geometric difference between the transformed source point cloud and target point cloud in the feature space, which can predict the alignment precision as the reward in planning.
%through feature difference, which is used to evaluate the alignment precision between the transformed source point cloud and target point cloud. 
%predict the new transformed state of the source point clouds while the evaluation network is used to evaluate the alignment precision between the transformed source point clouds and target point clouds. 
We formulate a transformation consistency loss on the transformation network and a geometry consistency loss on the evaluation network to train the dynamic model. Then, based on the learned model, we employ the cross-entropy method (CEM)~\cite{botev2013cross} to globally search for the optimal rigid transformation. CEM is a sampling-based planning algorithm, which alternately samples multiple transformations from a distribution and selects some  ``elite"  ones with the high evaluation scores to update the sampling distribution. 
The sampled transformations in each iteration are expected to approach to the optimal solution progressively and the mean of the elite transformations in the last iteration is viewed as the final solution, i.e., the transformation between the source and target point clouds. 
Experimental results on benchmark datasets demonstrate that the proposed registration algorithm can yield good performance. 

In summary, the contributions of this paper are as follows: 
\begin{itemize}
	\item We model the point cloud registration task as a Markov decision process and propose a latent dynamic model of point clouds for planning.
	\item We perform planning with the learned latent dynamic model for unsupervised point cloud registration.
	\item Extensive experiments on the standard benchmark datasets verify the outstanding registration performance of the proposed method.
\end{itemize}

\section{Related Work}
\paragraph{Traditional point cloud registration algorithms.}  
Iterative closest point (ICP) algorithm \cite{besl1992method} is a popular method for fine registration, which alternates between estimating the correspondences and updating the transformation based on the least squares optimization. 
However, with an improper initial transformation, it easily falls into the local minima.
To relieve it, Go-ICP \cite{yang2013go} is proposed to search for the optimal transformation by combining ICP with the branch-and-bound scheme.
%Other variants such as \cite{gressin2013towards,deng2018ppfnet} also exhibit competitive registration performance.
In addition, RANSAC based methods \cite{fischler1981random,chum2005matching} are also widely-used for coarse registration where wide bases are sampled for aligning the point cloud pair. 4PCS \cite{aiger20084} is a representative RANSAC based method, which proposes to sample four points with an approximately co-planar layout as a base set and identify the corresponding sets in the point cloud pair by comparing the intersectional diagonal ratios of the base set. 
Moreover, other RANSAC variant \cite{mellado2014super} also shows good performance.

\paragraph{Learning based point cloud registration algorithms.} 
In recent years, more and more attention has been paid to deep learning based point cloud registration. 
Benefitting from that PointNet \cite{qi2017pointnet} can directly extract robust feature of unordered point cloud with deep neural network, this research direction has obtained great progress.
PRNet \cite{wang2019prnet}  is an iterative method that utilizes the keypoint detection to handle the partially overlapping point cloud pair in a self-supervised way. 
%Moreover, Deng \textit{et al.} \cite{deng20193d} proposed to directly estimate the rigid transformation with a multi-layer perceptron (MLP) based on the extracted feature from FoldingNet \cite{yang2018foldingnet}. 
Instead, DCP \cite{wang2019deep} performs the SVD module to estimate the transformation based on a soft matching between the point cloud pair.
RPM-Net \cite{yew2020rpm} also realizes the registration based on the matching map generated from the differentiable Sinkhorn layer and annealing.
%DeepGMR \cite{yuan2020deepgmr} aligns the point cloud pair through minimizing the KL-Divergence between two learned Gaussian mixture models.
Yang \textit{et al.} \cite{yang2020mapping} proposed an unsupervised learning method for finding the correspondences based on the cycle consistency.
%Li \textit{et al.} \cite{li2020unsupervised} aimed to exploit the shape completion to handle the partial case. 
In addition, other learning-based methods including \cite{aoki2019pointnetlk,li2020unsupervised} also exhibit impressive performance.

\begin{figure*}
	\centering
	\begin{subfigure}[b]{1.0\linewidth}
		\includegraphics[width=\linewidth]{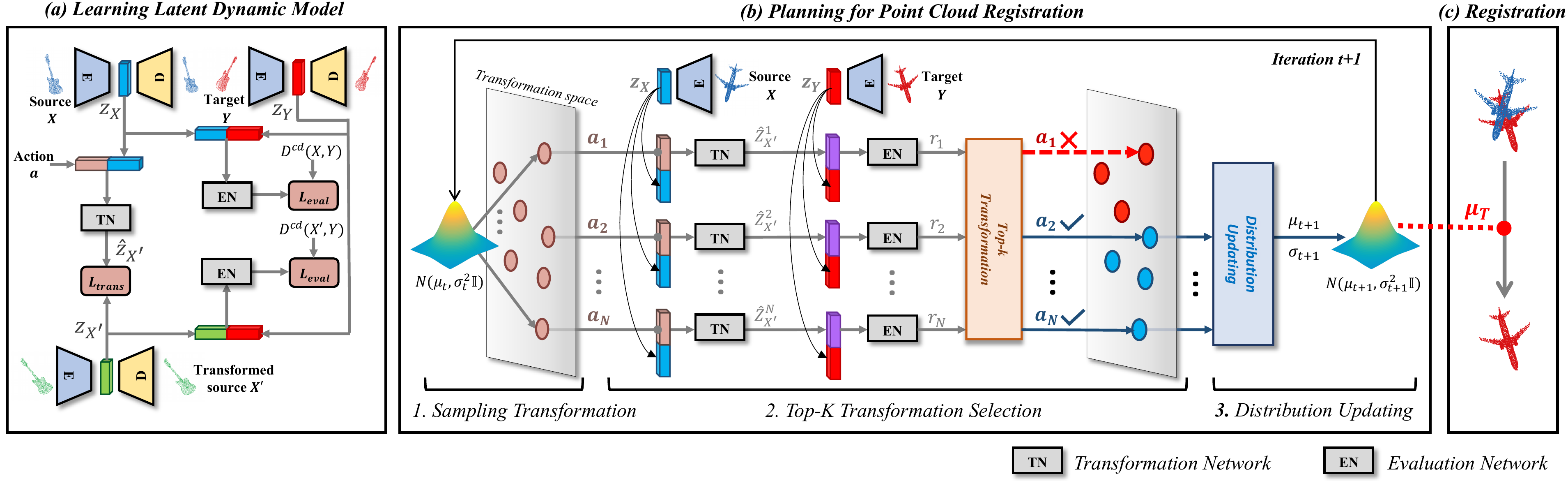}
	\end{subfigure}
	\caption{The pipeline of our proposed planning based point cloud registration method. (a) The left block demonstrates the framework of our latent dynamic model learning. For each generated sample $\langle\mathbf{X}, \mathbf{a}, \mathbf{X}', \mathbf{Y}\rangle$ as in Section 3.2, 
		%including the source and target point clouds $\mathbf{X}$ and $\mathbf{Y}$, the sampled transformation $\mathbf{a}$ and the transformed point cloud $\mathbf{X}'$ after performing $\mathbf{a}$ on $\mathbf{X}$, 
		the encoder first maps the point clouds $\mathbf{X}$ , $\mathbf{X}'$ and $\mathbf{Y}$ into the latent states $\mathbf{z}_X$, $\mathbf{z}_{X'}$ and $\mathbf{z}_Y$. Then, the concatenated vector of latent state $\mathbf{z}_X$ and action $\mathbf{a}$ is fed into the latent transformation model to predict the next latent state $\hat{\mathbf{z}}_{X'}$, which is trained by transformation consistency loss $\mathcal{L}_{trans}$ (see Eq.~\ref{trans}). Furthermore, for inputs $\mathbf{z}_X$ and $\mathbf{z}_Y$, the evaluation function aims to predict the Chamfer distance between $\mathbf{X}$ and $\mathbf{Y}$ by optimizing the geometric consistency loss $\mathcal{L}_{eval}$ (see Eq.~\ref{eval}). (b) The middle block shows the planning process with the learned latent dynamic model, which performs the steps: randomly sampling $N$ transformation candidates $\left\{\mathbf{a}_1,\mathbf{a}_2,\ldots,\mathbf{a}_N\right\}\sim\mathcal{N}\left(\mathbf{\mu}_t, \mathbf{\sigma}_t^2\mathbb{I}\right)$;  evaluating the reward $r_i$ of each transformation $\mathbf{a}_i$: $r_i=r\left(\mathbf{z}_X, \mathbf{a}_i,\mathbf{z}_Y\right)=-d_{\psi}\left(\left[f_\theta^z\left(\left[\mathbf{z}_X, \mathbf{a}_i\right]\right),  \mathbf{z}_Y\right]\right)$; choosing the top-$K$ transformations with the highest rewards; updating the sampling distribution with the top-K transformations via Eq.~\ref{update}.
%First, we encode the source and target point clouds $\mathbf{X}$ and $\mathbf{Y}$ into latent states: $\mathbf{z}_X=\varphi^{enc}\left(\mathbf{X}\right)$ and $\mathbf{z}_Y=\varphi^{enc}\left(\mathbf{Y}\right)$. Then, during the $t$-th iteration ($0\leq t\leq T-1$), we first randomly sample $N$ transformation candidates $\left\{\mathbf{a}_1,\mathbf{a}_2,\ldots,\mathbf{a}_N\right\}\sim\mathcal{N}\left(\mathbf{\mu}_t, \mathbf{\sigma}_t^2\mathbb{I}\right)$; Then, we evaluate the reward $r_i$ of each candidate $\mathbf{a}_i$: $r_i=r\left(\mathbf{z}_X, \mathbf{a}_i,\mathbf{z}_Y\right)=-d_{\psi}\left(\left[f_\theta^z\left(\left[\mathbf{z}_X, \mathbf{a}_i\right]\right),  \mathbf{z}_Y\right]\right)$; Further, we choose top-$K$ transformations with the highest rewards; Finally, the top-$K$ transformations are used to update the sampling distribution via Eq.~\ref{update} and the updated distribition will be used in next iteration. 
		(c) The right block shows that we perform transformation $\mu_T$ on the source point cloud $\mathbf{X}$ to align the target $\mathbf{Y}$.}
	\label{pipeline2}
\end{figure*}

%by planning with the learned latent dynamic model. First, we encode the given source and target point clouds $\mathbf{X}$ and $\mathbf{Y}$ into latent states: $\mathbf{z}_X=\varphi^{enc}\left(\mathbf{X}\right)$ and $\mathbf{z}_Y=\varphi^{enc}\left(\mathbf{Y}\right)$. Then, we perform cross entropy method with a learned dynamic model for finding the optimal solution. In detail, during $t$-th iteration, (1) we randomly sample $N$ transformation candidates $\left\{\mathbf{a}_1,\mathbf{a}_2,\ldots,\mathbf{a}_N\right\}\sim\mathcal{N}\left(\mathbf{\mu}_t, \mathbf{\sigma}_t^2\mathbb{I}\right)$; (2) We evaluate the reward $r_i$ of each candidate $\mathbf{a}_i$: $r_i=r\left(\mathbf{a}_i\mid\mathbf{z}_X,\mathbf{z}_Y\right)=-d_{\psi}\left(\left[f_\theta^z\left(\left[\mathbf{z}_X, \mathbf{a}_i\right]\right),  \mathbf{z}_Y\right]\right)=-d_{\psi}\left(\left[\hat{\mathbf{z}}_{X'}^i,  \mathbf{z}_Y\right]\right)$; (3) We choose $K$ candidates with the highest rewards as the elite candidates; (4) The elite candidates are exploited to update the sampling distribution via Eq.~\ref{update} and the new $\mathcal{N}\left(\mathbf{\mu}_{t+1}, \mathbf{\sigma}_{t+1}^2\mathbb{I}\right)$ will be used in next iteration. Finally, we perform rigid transformation $\mathbf{\mu}_T$ on source point cloud $\mathbf{X}$ to align the target $\mathbf{Y}$.

\paragraph{Planning based optimal decision.} Planning based decision algorithm is important and well-studied in the optimal control field. Typically, those methods search for the optimal action via repeatedly performing forward simulation and future state evaluation. For discrete control task, Monte-Carlo tree search (MCTS) \cite{browne2012survey} is one of the most widely used method whose each simulation procedure contains selection, evaluation, expansion and backing-propagation. To handle exploration-exploitation dilemma in MCTS, UCT and its variants~\cite{coulom2006efficient,rosin2011multi} add an additional exploration bonus on original estimated reward for each action. 
%Furthermore, various parallel versions of MCTS \cite{chaslot2008parallel,segal2010scalability,enzenberger2009lock} are proposed to improve the construction speed of the search tree.
For the continuous action task, the cross entropy method (CEM)~\cite{botev2013cross} is a powerful planning method, which can obtain the optimal action via gradually narrowing the searching space in a trial-and-error way.
To improve the robustness of CEM, Liu \textit{et al.}~\cite{liu2020safe} proposed to perform CEM with model uncertainty and constraints. 
%Furthermore, Pinneri \textit{et al.} improved CEM in terms of the sampling efficiency so that it can make the faster decision. 
Furthermore, the differentiable CEM~\cite{amos2020differentiable} is proposed to allow the end-to-end training.
 %Bharadhwaj \textit{et al.}~\cite{bharadhwaj2020model} combined the CEM with the gradient based optimization to improve the decision performance. 
 Other variants or applications~\cite{moss2020cross,hafner2019learning} also show its great potential.

\section{Approach}
In this section, we present the proposed point cloud registration method by planning with the latent dynamic model. Fig.~\ref{pipeline2} illustrates the proposed framework, which consists of two components: learning latent dynamic model of point clouds and planning for point cloud registration. In Section 3.1, we define the point cloud registration problem with the Markov decision process. In Sections 3.2 and 3.3, we present the details of learning the latent dynamic model  and planning for point cloud registration, respectively.
\subsection{Problem Setting}
Given the source point cloud $\mathbf{X}=\{\mathbf{x}_{1},\mathbf{x}_{2},\ldots \mathbf{x}_{N}\}$ and target point cloud $\mathbf{Y}=\{\mathbf{y}_{1}, \mathbf{y}_{2},\ldots \mathbf{y}_{N}\}$, where $\mathbf{x}_{i} \in \mathbb{R}^{3}$ and $\mathbf{y}_{i} \in \mathbb{R}^{3}$, $i=1,2,\ldots,N$, the point cloud registration task aims to find the rigid transformation between $\mathbf{X}$ and $\mathbf{Y}$, containing the rotation $\mathbf{R}\in SO(3)$ and the translation $\mathbf{t}\in \mathbb{R}^{3}$. 
The rigid transformation in 3D Euclidean space is defined as $\mathbf{X}'=f^{3d}\left(\mathbf{X},\mathbf{R},\mathbf{t}\right)=\mathbf{R} \cdot \mathbf{X}^{\top}+\mathbf{t}$.
Formally, we model the point cloud registration problem as a Markov decision process in the feature space of point clouds, which is defined by a latent state space, an action space, a latent transformation function and a latent reward function.
\paragraph{Latent state space.} The latent state space of point clouds $\mathcal{Z} \subseteq \mathbb{R}^{d_{n}}$ is defined as a $d_n$-dimensional compact embedding space. Given the point cloud $\mathbf{X}$, we employ an encoding network $\varphi_{\phi}^{\rm enc}$ to map $\mathbf{X}$ into its corresponding latent state $\mathbf{z}_{X} =\varphi_{\phi}^{\rm enc}\left(\mathbf{{X}}\right) \in \mathcal{Z}$.

\paragraph{Action space.} We define the rigid transformation as the action. 
The action space $\mathcal{A}$ is the rigid transformation space of point clouds $SE(3)$  ($SO(3) \times \mathbb{R}^{3}$). 
To reduce the dimension of the search space, we use the euler-angle representation $\mathbf{e}=[e_1,e_2,e_3] \in [-\pi, \pi]^3$  to encode the rotation $\mathbf{R} \in SO(3)$. 
Therefore, the action $\mathbf{a} \in \mathcal{A}$ is represented by $[\mathbf{e}, \mathbf{t}]$ and  $\mathbf{R}(\mathbf{e})$ is the corresponding rigid rotation matrix of euler-angle representation $\mathbf{e}$. 

\paragraph{Latent transformation function.} 
In the 3D Euclidean space, after performing the action $\mathbf{a}$, the point cloud $\mathbf{X}$ will be transformed to $\mathbf{X}'$ via the rigid transformation. 
Similarly, in the latent state space, the latent transformation function $f_\theta^z: \mathcal{Z} \times \mathcal{A} \rightarrow \mathcal{Z}$ aims to infer the next latent state $\hat{\mathbf{z}}_{X'}$ by performing action $\mathbf{a}$ at the state $\mathbf{z}_X$. It is defined as below:
\begin{equation}
\hat{\mathbf{z}}_{X'}=f_\theta^z\left(\left[\mathbf{z}_X, \mathbf{a}\right]\right),
\end{equation}
where $\left[\cdot,\cdot\right]$ denotes the concatenation operation. 
Since the transformation function in the registration task is nonlinear, we utilize a deep network with parameters ${\theta}$ to model it.

\paragraph{Latent reward function.} 
For the source point cloud $\mathbf{X}$ and target point cloud $\mathbf{Y}$, given their latent states $\mathbf{z}_X$ and $\mathbf{z}_Y$,  we use the output of an evaluation network as the reward signal after executing the action $\mathbf{a}$ at the latent state $\mathbf{z}_X$.  Specifically, the latent reward function $r: \mathcal{Z}\times\mathcal{A}\times\mathcal{Z}\rightarrow\mathbb{R}$ is defined as:
\begin{equation}
\begin{split}
r\left(\mathbf{z}_X,\mathbf{a},\mathbf{z}_Y\right) &=-d_{\psi}\left(\left[f_\theta^z\left(\left[\mathbf{z}_X,\mathbf{a}\right]\right), \mathbf{z}_Y\right]\right)\\
&=-d_{\psi}\left(\left[\hat{\mathbf{z}}_{X'}, \mathbf{z}_Y\right]\right),
\end{split}
\end{equation}
where the evaluation network $d_{\psi}:\mathcal{Z}\times\mathcal{Z}\rightarrow\mathbb{R}$ estimates the alignment precision between $\mathbf{X}'$ and $\mathbf{Y}$ in the feature space. The point cloud registration task aims to find the optimal policy $\mathbf{a}^*$ that can maximize the reward:
\begin{equation}
\label{target}
\begin{aligned} 
\mathbf{a}^*&=\arg\max _{\mathbf{a}\in \mathcal{A}} r\left( \mathbf{z}_X, \mathbf{a}, \mathbf{z}_Y\right)\\
&= \arg\max _{\mathbf{a}\in \mathcal{A}} -d_{\psi}\left(\left[f_\theta^z\left(\left[\mathbf{z}_X, \mathbf{a}\right]\right),  \mathbf{z}_Y\right]\right).
\end{aligned}
\end{equation}

\subsection{Learning Latent Dynamic Model}
Our dynamic model of point clouds is constructed in the feature space of point clouds, where we employ an encoder-decoder network to extract deep features (latent states) of point clouds so that the local geometric structures of point clouds can be characterized well. 
As shown in Fig.~\ref{pipeline2} (a), by fusing the feature of the source point cloud and the executed action, we first form a transformation network to predict the latent state of the transformed source point cloud.
%By encoding the feature of the source point cloud and the executed action at the state with multi-layer perceptrons (MLPs), we first form a transformation network to predict the new transformed state of the source point clouds. 
%Thus, we can obtain the features of the transformed source point clouds. 
We then employ an evaluation network to evaluate the alignment precision between the transformed source point cloud and the target point cloud in the feature space. 
The evaluated alignment precision between point clouds is used as the reward in planning for point cloud registration. 
By performing a set of actions in trial and error to maximize the reward during the registration process, we can seek the action with the highest reward, i.e., the transformation between a pair of point clouds (please refer to Section 3.3).

Before effectively training our dynamic model, we need to generate training samples at first. For each pair of point clouds $\langle\mathbf{X}, \mathbf{Y}\rangle$, we sample multiple actions $\left\{\mathbf{a}=\left[\mathbf{e}, \mathbf{t}\right]\right\}$,  where each action sample $\mathbf{a}$ follows a normal distribution $\mathcal{N}\left( \mathbf{0}, \mathbf{\sigma}^{2} \mathbb{I} \right)$ over the action space $\mathcal{A}$. % and the standard deviation $\mathbf{\sigma}$ controls the sampling region.  
Next, we perform each sampled action on $\mathbf{{X}}$ and obtain the transformed point clouds $\{\mathbf{{X}}'\}$ where  $\mathbf{{X}}' = f^{3d}\left(\mathbf{{X}}, \mathbf{R}(\mathbf{e}), \mathbf{t} \right)$. Thus, we can obtain the generated training samples $\mathcal{S}=\left\{\mathbf{{X}}, \mathbf{a}, \mathbf{{X}}', \mathbf{{Y}}\right\}$.

During the model learning, we jointly train an auto-encoder and a latent dynamic model. 
The auto-encoder is expected to learn the discriminative feature representation (latent state) of point clouds. 
The encoder $\varphi_{\phi}^{enc}$ and decoder $\varphi_{\omega}^{dec}$ modules in the auto-encoder are trained via the reconstruction loss: 
\begin{equation}
\label{rec}
\mathcal{L}_{rec}\left(\phi,\omega\right) =\frac{1}{\left|\mathcal{S}\right|} \sum_{\langle\mathbf{X}, \mathbf{X}', \mathbf{Y}\rangle\in \mathcal{S}}\mathcal{L}\left(\mathbf{X}\right) + \mathcal{L}\left(\mathbf{X}'\right) + \mathcal{L}\left(\mathbf{Y}\right),
\end{equation}
where $\mathcal{L}\left(\mathbf{X}\right)={D^{cd}\left(\varphi_{\omega}^{\rm dec}\left(\varphi_{\phi}^{\rm enc}(\mathbf{{X}})\right), \mathbf{{X}}\right)}$. $D^{cd}$ is the Chamfer distance between $\mathbf{X}$ and $\mathbf{Y}$, $D^{cd}\left(\mathbf{X}, {\mathbf{Y}}\right)=$
\begin{equation}
\begin{split} \sum_{\mathbf{x}_{i} \in \mathbf{X}} \min _{\mathbf{\mathbf { y }}_{j} \in {\mathbf{Y}}}\left\|\mathbf{x}_{i}-{\mathbf{y}}_{j}\right\|_{2}+ \sum_{{\mathbf{y}}_{i} \in {\mathbf{Y}}} \min _{\mathbf{x}_{j} \in \mathbf{X}}\left\|{\mathbf{y}}_{i}-\mathbf{x}_{j}\right\|_{2}.
\end{split}
\end{equation}
Notably, once the encoder and decoder are trained, the encoder is only used for testing.

For the source point clouds $\mathbf{X}$ , the transformation network $f_\theta^z$ is expected to capture the feature changes when a point cloud is transformed by performing the action on them. That is the predicted transformed state $f_\theta^z\left(\mathbf{z}_X, \mathbf{a}\right)$  is as consistent as possible to the state $\mathbf{z}_{X'}$ of the transformed point clouds. 
%and thus  can predict the new transformed state $f_\theta^z\left(\mathbf{z}_X, \mathbf{a}\right)$ after executing an action at the current state $\mathbf{z}_X$. It is expected that  $f_\theta^z$ is sensitive to the feature changes when a point cloud is transformed and the transformed state is as consistent as possible to the state $\mathbf{z}_{X'}$ of the transformed point clouds by performing the action on the source point clouds $\mathbf{X}$. 
Therefore, we formulate a transformation consistency loss $\mathcal{L}_{trans}$ to train our transformation network:
\begin{equation}
\label{trans}
\mathcal{L}_{trans}\left(\theta\right)=\frac{1}{\left|\mathcal{S}\right|}\sum_{\langle\mathbf{X}, \mathbf{a}, \mathbf{X}'\rangle\in \mathcal{S}}{\|f_\theta^z\left(\left[\mathbf{z}_X,\mathbf{a}\right]\right)-\mathbf{z}_{X'}\|_2^2}.
\end{equation}

In our dynamic model, the evaluation network $d_\psi$ aims to predict the alignment precision between a pair of point clouds with their states, which can characterize the geometric differences of point clouds in the feature space. We encourage the geometric differences of point clouds in the feature and spatial spaces to be consistent. To this end, we propose a geometric consistency loss $\mathcal{L}_{eval}$ to train our network:
\begin{equation}
\label{eval}
\mathcal{L}_{eval}\left(\psi \right) = \frac{1}{|\mathcal{S}|} \sum_{\langle \mathbf{X}, \mathbf{{X}}', \mathbf{Y}\rangle\in\mathcal{S}}\mathcal{I}\left(\mathbf{X}, \mathbf{Y}\right) + \mathcal{I}\left(\mathbf{X}', \mathbf{Y}\right),
\end{equation}
where $\mathcal{I}\left(\mathbf{X}, \mathbf{Y}\right)=\left( d_{\psi}\left( \mathbf{z}_X, \mathbf{z}_{Y}\right) - D^{cd}\left(\mathbf{{X}}, \mathbf{{Y}}\right)\right)^{2}$, $d_{\psi}\left(\mathbf{z}_X, \mathbf{z}_Y \right)$ is the output of the evaluation network. Notably,  we utilize the Chamfer distance to quantify the alignment precision between a point clouds pair as in \cite{yew2020rpm}, since as a distance metric, it tends to obtain the lower value when the point clouds are better aligned.
%since as a distance metric, it tends to obtain lower distance value when measures a better aligned pair.
%we utilize the Chamfer distance to describe the alignment precision between the transformed source point cloud and target point cloud as \cite{yew2020rpm}.

Finally, we jointly train the encoder-decoder and latent dynamic model via the following loss function:
\begin{equation}
\mathcal{L}\left( \theta, \phi, \omega, \psi \right) = \mathcal{L}_{ rec}\left(\phi, \omega \right) + \mathcal{L}_{trans}\left(\theta\right) + \mathcal{L}_{eval}\left(\psi\right).
\end{equation}

\begin{algorithm}
	\caption{CEM based Registration with Learned Latent Dynamic Model}
	\label{alg:algorithm}
	\textbf{Input:} Encoder $\varphi_\phi^{enc}$, latent transformation network $f_\theta^z$, latent evaluation network $d_\psi$, iterations  $T$, sampled transformations $N$,  top transformations $K$, initial Gaussian distribution $\mathcal{N}\left(\mathbf{\mu}_0, \mathbf{\sigma}_0^2\mathbb{I}\right)$, source and target point clouds $\mathbf{X}$ and $\mathbf{Y}$.\\
	\textbf{Output:} Estimated rigid transformation between $\mathbf{X}$ and $\mathbf{Y}$.
	
	\begin{algorithmic}[0] %[1] enables line numbers
		\STATE Encode $\mathbf{X}$ and  $\mathbf{Y}$: $\mathbf{z}_X=\varphi_{\phi}^{enc}(\mathbf{{X}})$ and  $\mathbf{z}_Y=\varphi_{\phi}^{enc}(\mathbf{{Y}})$.
		\FOR{$t = 0:T-1$}
		\FOR{$i = 1:N$}
		\STATE Sample transformation $\mathbf{a}_t^{(i)}\sim \mathcal{N}\left(\mathbf{\mu}_t, \mathbf{\sigma}_t^2\mathbb{I}\right)$.
		\STATE Perform transformation: $\hat{\mathbf{z}}_{X'}^{i}=f_\theta^z\left(\left[\mathbf{z}_X, \mathbf{a}\right]\right)$.
		\STATE Calculate reward $r_i=-d_\psi\left(\left[\hat{\mathbf{z}}_{X'}^i, \mathbf{z}_{Y}\right]\right)$.
		\ENDFOR
		\STATE Sort the rewards $\left\{r_i\right\}$ and select top transformations $\mathcal{K}$ with $K$ largest rewards.
		\STATE Update the sampling distribution with $\mathcal{K}$ via Eq.~\ref{update}.
		\ENDFOR
		
		\textbf{Return:} $\mathbf{\mu}_T$
	\end{algorithmic}
	\label{alg1}
\end{algorithm}

\subsection{Planning for Point Cloud Registration}
As shown in Fig.~\ref{pipeline2} (b), once the dynamic model of point clouds is trained, given the source and target point clouds, by performing actions on the source point clouds, we can maximize the rewards (i.e., the outputs of the evaluation network) to obtain the optimal action for registration. In this subsection, we employ the cross entropy method (CEM) to obtain the desired transformation for registration. CEM is an efficient sampling-based planning algorithm that searches for the optimal solution via iteratively narrowing the searching region in a trial-and-error way. As outlined in Algorithm 1, in the $t$-th iteration ($0\leq t<T-1$), it contains three steps:
\paragraph{(a) Random transformation sampling.} We randomly sample $N$ rigid transformations $\{\mathbf{a}_t^{(1)}, \ldots, \mathbf{a}_t^{(N)}\}$ from a parameterized Gaussian distribution $\mathcal{N}\left( \mathbf{\mu}_t, \mathbf{\sigma}_t^2\mathbb{I} \right)$ over the action space $\mathcal{A}$.
 Since the dimension of the action space is 6, the mean $\mathbf{\mu}_t$ and variance $\mathbf{\sigma}_t^2$ are represented by  $\left( \mu_{t,1},\ldots, \mu_{t,6} \right)$ and $\left( \sigma_{t, 1}^2, \ldots, \sigma_{t, 6}^2 \right)$, respectively.
 
 \paragraph{(b) Top-$K$ transformation selection.} 
 From all sampled $N$ rigid transformations, with the trained transformation network and evaluation network, we choose top-$K$ ones with the highest rewards, denoted as  $\mathcal{K}=\{\mathbf{a}_t^{(i)} \mid r\left(\mathbf{z}_X, \mathbf{a}_t^{(i)}, \mathbf{z}_Y\right)=-d_{\psi}\left(\left[f_\theta^z\left(\left[\mathbf{z}_X, \mathbf{a}\right]\right),  \mathbf{z}_Y\right]\right) \geq \beta_{t},$$ 1 \leq i \leq N\}$, where $\beta_t$ is the $K$-th highest reward.

 \paragraph{(c) Distribution updating.} 
 This step aims to update the parameters of the current sampling distribution so that in the next iteration we can sample ``better'' transformations from the updated distribution $\mathcal{N}\left(\mathbf{\mu}_{t+1}, \mathbf{\sigma}_{t+1}^{2} \mathbb{I}\right)$:
\begin{equation}
\begin{split}	
\mu_{t+1, j}=\frac{1}{K} \sum_{\mathbf{a} \in \mathcal{K}} a_{j},\ \sigma_{t+1, j}^{2}=\frac{1}{K} \sum_{\mathbf{a} \in \mathcal{K}}\left(a_{j}-\mu_{t+1, j}\right)^{2},
\end{split}
\label{update}
\end{equation}
where $j=1, \ldots, 6$. 
\begin{table*}[t]
	\centering
	\resizebox{1.0\linewidth}{!}{
		\begin{tabular}{lcccc|cccc|cccc}
			\toprule
			\multirow{2}{*}{{Model}} & \multicolumn{4}{c|}{{ModelNet40-Same categories}} & \multicolumn{4}{c|}{{ModelNet40-Gaussian noise}} & \multicolumn{4}{c}{{7Scene}} \\
			& {RMSE(R)}& {RMSE(t)}& {MAE(R)} & {MAE(t)} & {RMSE(R)}& {RMSE(t)}& {MAE(R)} & {MAE(t)} &  {RMSE(R)}& {RMSE(t)}& {MAE(R)} & {MAE(t)}            \\ \midrule 
			ICP & 11.6165 & 0.0255 & 2.5852 & 0.00715 & 11.3945 & 0.0218 & 2.4573 & 0.0041 & 21.0034 & 0.0262 & 5.156972 & 0.005913 \\ 
			Go-ICP & 11.8523 & 0.0256 & 2.5884 & 0.00709 & 11.4534 & 0.0230 &  2.5348 & 0.0041 & 5.4590 & 0.0050 & 0.684649  & 0.000659  \\ 
			Super4PCS &4.6191 & 0.0050 & \textbf{0.2024} & 0.00038 & 19.6369 & 0.0980 & 3.9660 & 0.0175 &13.4823 & 0.0101 & 1.410883 & 0.001253 \\ 
			FGR & 4.1355 &0.0076 & 0.3161 & 0.00060 &  24.6514 & 0.1089 & 10.0559 & 0.0273 & 0.2756 & \underline{0.0026} & 0.120562 & 0.000637 \\
			PointNetLK & 5.1963 & 0.0051 & 1.0231 & 0.00057 &5.2422 & 0.0080 & 1.5974 & {0.0053} &  0.0544 & \textbf{0.0007} & 0.006484 & 0.000007 \\ 
			FMR & 10.9514 & 0.0113 & 3.8465 & 0.00396 &11.1682 & 0.0124 & 4.2535 & 0.0062&7.5498 & 0.0133 & 1.995638 & 0.003718 \\ 
			IDAM & \textbf{1.5093} & 0.0247 & 1.0318 & 0.01686 &3.1485 & 0.0109 & 0.9739 & 0.0048 & 7.8643 & 0.0181 & 1.757570 & 0.007679 \\
			%			RPMNet ~\cite{yew2020rpm} $(\lozenge)$ & \textbf{0.4794} & 0.0021 & \textbf{0.0309} & {0.0003}&\textbf{0.7882}&0.0047&\textbf{0.0523}&0.0006&1.6077 & 0.0117 & 0.4756 & 0.0040 \\
			\midrule
			LatentCEM & \underline{3.0178} & \textbf{0.0028} & \underline{0.2779} & \textbf{0.00036} & \textbf{3.0681} & \textbf{0.0066} & \textbf{0.5853} & \textbf{0.0016} &\textbf{0.0066} & {0.0030} & \textbf{0.000015} & \textbf{0.000004} \\ \bottomrule
	\end{tabular}}
	%	\vspace{0.5mm
	\caption{Comparison results on {ModelNet40-Same categories}, {ModelNet40-Gaussian noise} and 7Scene datasets.}\label{tab:modelnet40}
	%	\vspace{-5mm}
\end{table*}
In fact, Eq.~\ref{update} is the closed-form solution of the maximum-likelihood estimation problem over the top-$K$ transformations, that is $\mathbf{ \mu}_{t+1},\mathbf{ \sigma}_{t+1}=\arg\max_{\mathbf{\mu},\mathbf{\sigma}}\sum_{\mathbf{a}\in\mathcal{K}}g\left(\mathbf{a};\mathbf{\mu},\mathbf{\sigma}\right)$, where $g\left(\cdot;\mathbf{\mu}, \mathbf{\sigma} \right)$ is the probability density function of $\mathcal{N}\left(\mathbf{\mu},\mathbf{\sigma}^2\right)$. By re-fitting the sampling distribution to the top-$K$ samples, the updated sampling distribution tends to focus on more promising transformation region and the sampled transformations in the next iteration are expected to obtain the higher rewards. 
In the last iteration, we utilize the mean $\mathbf{\mu}_T$ of the distribution to estimate the optimal solution.

\begin{figure*}[htbp] \centering 	
	\includegraphics[width=1\linewidth]{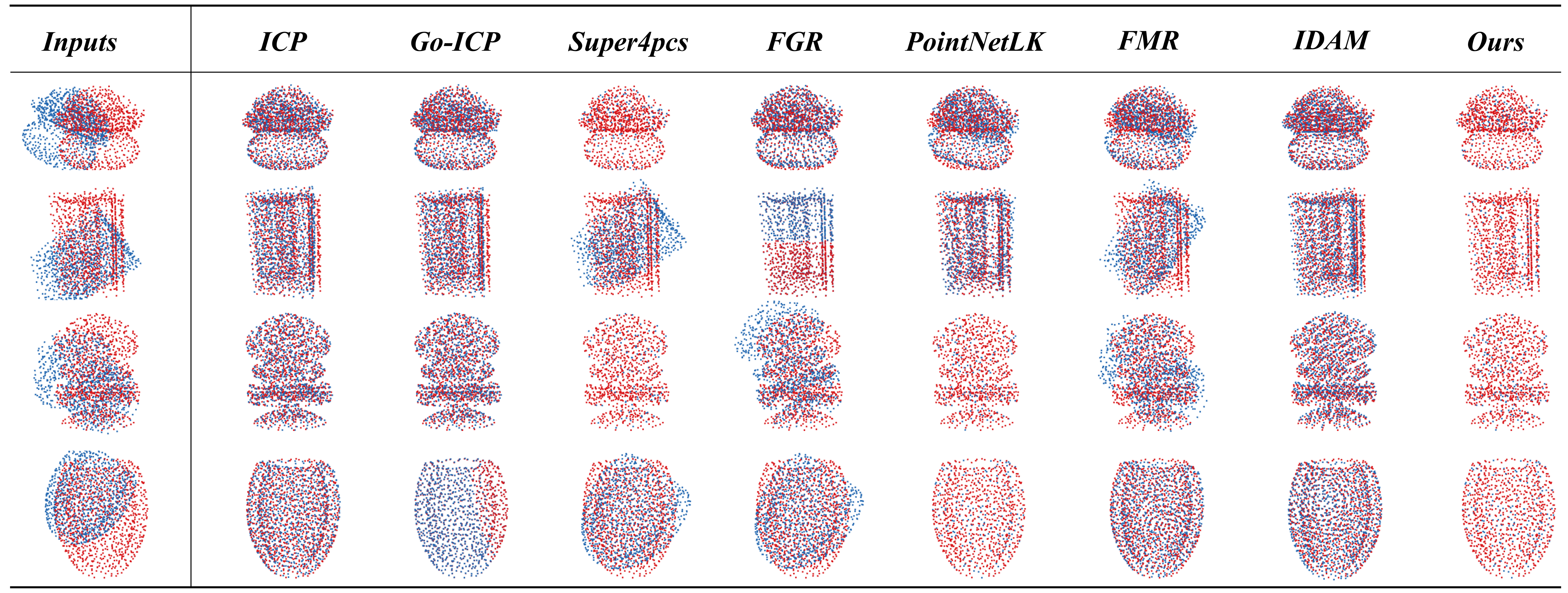} 
	\caption{The qualitative comparison with ICP, Go-ICP, Super4PCS, FGR, PointNetLK, FMR and IDAM on some example point clouds.} 
	\label{visual} 
\end{figure*} 

\section{Experiments}
%In this section, we first present the implementation details of our proposed method. Then, we compare our proposed method with some state-of-the-art methods on benchmark datasets containing ModelNet40~\cite{wu20153d} and 7Scene~\cite{shotton2013scene}. Finally, we perform ablation studies to further verify the effectiveness of our algorithm. For simplicity, we denote our CEM based latent planning method for registration as \textbf{LatentCEM}.

\subsection{Implementation Details}
For feature extraction of point clouds, we employ the DGCNN~\cite{wang2019dynamic} as encoder and the decoder of FoldingNet~\cite{yang2018foldingnet} as decoder in our auto-encoder. 
The latent transformation and evaluation networks are modeled by two 3-layer MLPs and more details are described in Appendix 2.
We use the output of the encoder as the latent state of the point cloud whose dimension is set to 512.
For CEM, the numbers of iterations $T$, candidates $N$ and elite candidates $K$ in CEM are set to 10, 1000 and 25, respectively. 
The $\boldsymbol{\sigma}$ used in training samples generation in Section 3.2 is set to $\boldsymbol{1}$.
Our method is implemented in PyTorch and the model is trained with the Adam optimizer , where the learning rate and weight decay are set to $0.0001$ and $0.0005$. For simplicity, we denote our CEM based latent planning method for registration as \textbf{LatentCEM}.

\subsection{Evaluation on ModelNet40}
We perform performance comparison with four traditional registration methods including ICP~\cite{besl1992method}, Go-ICP~\cite{yang2013go} , Super4PCS~\cite{mellado2014super} and FGR~\cite{zhou2016fast}, and three deep learning methods including fully-supervised PointNetLK~\cite{aoki2019pointnetlk}, fully-supervised IDAM~\cite{li2019iterative} and unsupervised FMR~\cite{huang2020feature} on the ModelNet40 dataset. The parameter settings of the traditional methods can be seen in Appendix 3.
The ModelNet40 dataset contains $12,311$ mesh CAD models from 40 object categories.
We select $2,468$ models for testing and uniformly sample $784$ points from the outer surface of each model, where selected points in each model are centered and rescaled to fit within the unit sphere. 
%The selected points in each model are centered and rescaled to fit within the unit sphere. 
%The $(x, y, z)$ coordinates are only used as the input. 
We utilize the root mean square error (RMSE) and mean absolute error (MAE) between ground truth and predicted transformations measure the quality of registration.
All angular error results are converted to the units of degrees. 
The target point clouds are obtained by transforming the source point clouds via a randomly generated transformation matrix. The rotation and translation along each axis uniformly lie in the $[0, 45^{\circ}]$ and $[-0.5, 0.5]$. 
Notably, no ground-truth transformation is involved in our model training and thus our method is unsupervised. Fig.~\ref{visual} shows the qualitative comparison  and more results can be seen in Appendix 1.

\textbf{Train and test on the same categories.} In this experiments, 
the samples for training our model and other three deep learning methods are from the same categories as the testing samples. Notably,  and the groud-truth transformations are required for PointNetLK and IDAM. As shown in Table~\ref{tab:modelnet40}, benefitting from the learned precise dynamic model and the effective CEM based search policy, our method can obtain the highest precision in terms of the translation error while the second highest precision in terms of the rotation error. And, although the RMSE of our method in rotation is worse than that of IDAM, the IDAM is fully-supervised, which needs amounts of transformation labels as the supervision signal while our method is unsupervised and thus can largely reduce the training cost.
%Further, Fig.~\ref{visual} and  Fig.~\ref{planning} show the qualitative comparison and  the planning process on some examples, respectively, and more visualization  results can be seen in Appendix 1.

%\begin{table}[htbp]
%	\caption{Comparison results on 7Scene.}
%	\label{modelnet1}
%	\centering
%	\label{7scene}
%	\resizebox{1.0\linewidth}{!}{
%		\begin{tabular}{l|cccc}
%			\toprule
%			\multirow{2}{*}{Model} & \multicolumn{2}{c}{RMSE} & \multicolumn{2}{c}{MAE} \\
%			& Rot.            & Trans.           & Rot.         & Trans.            \\ \midrule 
%			ICP~\cite{besl1992method}  & 21.003425 & 0.026211 & 5.156972 & 0.005913 \\
%			Go-ICP~\cite{yang2013go}  & 5.459007 & 0.005099 & 0.684649  & 0.000659 \\
%			Super4PCS~\cite{mellado2014super} & 13.482336 & 0.010198 & 1.410883 & 0.001253  \\
%			FGR~\cite{zhou2016fast} & 0.275646 & {0.002645} & 0.120562 & 0.000637 \\
%			PointNetLK~\cite{aoki2019pointnetlk} & 0.054424 & \textbf{0.000707} & 0.006484 & 0.000007 \\			
%			%SDRSAC~\cite{le2019sdrsac} &  &  &  & \\ 
%			FMR~\cite{huang2020feature} & 7.549876 & 0.013342 & 1.995638 & 0.003718 \\ 
%			IDAM~\cite{li2019iterative} & 7.864313 & 0.018193 & 1.757570 & 0.007679 \\ \midrule
%			LatentCEM (ours) & \textbf{0.006557} & {0.003000} & \textbf{0.000015} & \textbf{0.000004} \\ \bottomrule
%	\end{tabular}}
%\end{table}

\textbf{Gaussian noise.} 
To verify the robustness of our proposed point cloud registration method, we train our model on the noise-free data of ModelNet40 and test it on a new test dataset disturbed by Gaussian noise. 
Following \cite{wang2019prnet}, the noisy data is randomly sampled from a Gaussian distribution with mean $0$ and standard deviation $0.01$, and then clipped to $[-0.05, 0.05]$. Table~\ref{tab:modelnet40} shows that our method is robust to the noise and can obtain superior score than other methods. 

\subsection{Evaluation on 7Scene}
We further evaluate our method on the 7Scene dataset, which is a real indoor dataset containing seven scenes: Chess, Fires, Heads, Office, Pumpkin, Redkitchen and Stairs. Following \cite{huang2020feature}, We project the depth images into point clouds, and generate 296 scans for training and 57 scans for testing through the similar data processing in ModelNet40 dataset. Here, the target point cloud is also generated via performing a randomly generated transformation on the source point cloud.
Table~\ref{tab:modelnet40} summarizes the comparison results with other methods on the 7Scene dataset and our method can still yield good performance compared to other methods.

\subsection{Ablation Study and Analysis}
\paragraph{Encoder and decoder setting.}
For our auto-encoding framework, we conduct ablation studies on ModelNet40 with different encoder and decoder settings. The tested encoders include DGCNN and PointNet-like encoder with skip-links used in FoldingNet (named Folding\_Enc). 
%The former learns local geometric features via constructing the $k$-NN graph while the latter learns a global descriptor of the whole shape. 
The tested decoders include a MLP with three layers and the decoder of FoldingNet (named Folding\_Dec) which uses two consecutive $3$-layer MLPs to wrap a fixed 2D grid into the shape of the input point clouds. 
As shown in Table~\ref{autoencoder}, compared to other settings, benefiting from the local geometric features via constructing the $k$-NN graph in DGCNN and the effective reconstruction strategy in Folding\_Dec, the DGCNN and Folding\_Dec can yield better performance. 
%Therefore, we choose DGCNN+Folding\_Decoder as the auto-encoding module in our experiments.

\begin{table}[]
	\centering
	\label{module}
	\resizebox{1.0\linewidth}{!}{
		\begin{tabular}{l|cccc}
			\toprule
		 \textit{Model}	& {RMSE(R)}& {RMSE(t)}& {MAE(R)} & {MAE(t)} \\ \midrule 
			Folding\_Enc + MLP  & 4.35 & 0.0039 & 0.68 & 0.0007 \\
			Folding\_Enc + Folding\_Dec  & 3.58 & 0.0033 & 0.47 & 0.0005 \\
			DGCNN + MLP & 3.33 & 0.0036 & 0.43 & 0.0006 \\
			DGCNN + Folding\_Dec & {3.02} & {0.0028} & {0.28} & {0.0004} \\ \bottomrule
	\end{tabular}}
	\caption{Ablation study: Encoder and decoder setting.}
		\label{autoencoder}
\end{table}

\begin{figure}[htbp]
	\centering
	\begin{subfigure}[b]{1\linewidth}
		\includegraphics[width=\linewidth]{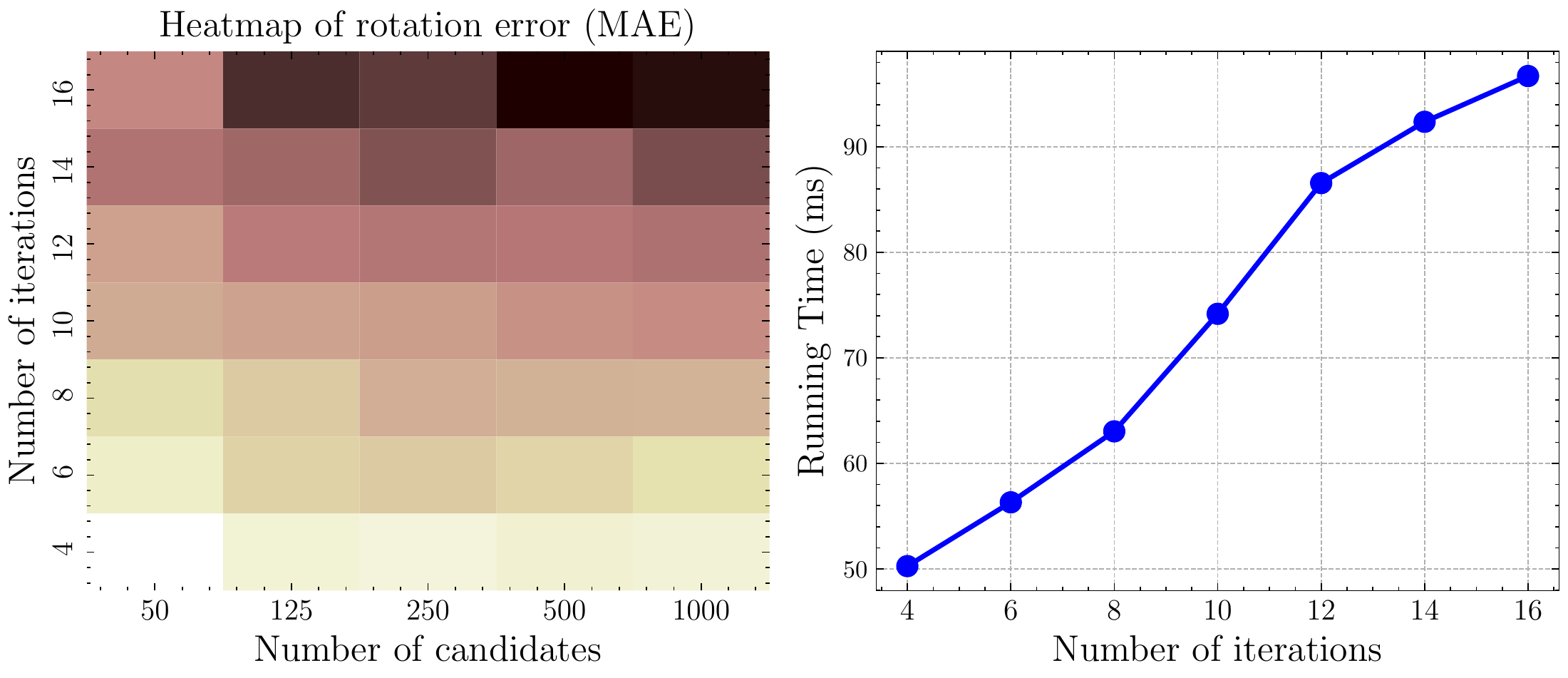}
		%\caption{More coffee.}
	\end{subfigure}
	\caption{
		The left heatmap plots rotation errors on ModelNet40 with different numbers of iterations $T$ and candidates $N$. The right figure shows the running time in the cases of different iterations.}
	\label{parameter}
\end{figure}

\paragraph{Parameters of CEM.}
We further analyze the sensitivity of hyper-parameters in CEM on registration precision, including the iteration number and candidate number.
%The heatmap plotted in Fig.~\ref{parameter} demonstrates the changes of rotation errors with different numbers of iterations $T=\left\{4, 6, 8, 10, 12, 14, 16\right\}$ and candidates $N=\left\{50, 125, 250, 500, 1000\right\}$.
%The heatmap plotted in Fig.~\ref{parameter} demonstrates the changes of rotation errors with different numbers of iterations $T$ and candidates $N$.
As shown in Fig.~\ref{parameter},  the iteration number and candidate number increase, the rotation errors tend to decrease (from blocks with light colors to blocks with deep colors).

\paragraph{Computational efficiency.} 
We also compare the inference time of different methods on the ModelNet40 dataset. 
The running time is measured in millisecond with a batch size of 1, averaged over the entire test set on a desktop computer with an Intel I5-8400 CPU and Geforce RTX 2080Ti GPU.
Notably, the traditional methods are executed on CPU  while deep learning methods on GPU.
The inference time of our method is $75.638$ and the inference time of other methods are  $8.168$ (ICP), $2739.588$  (Go-ICP), $6128.691$ (Super4PCS), $52.203$ (FGR), $386.244$ (FMR), $180.645$ (PointNetLK) and $41.401$ (IDAM), respectively. Fig.~\ref{parameter} shows the running time of our method in the cases of different iterations. From this figure, one can see that although our method can obtain higher registration precision with the increase of iteration times, it will need longer inference time. Therefore, it is necessary to perform good balance between the precision and the speed.

%\begin{figure}[htbp]
%	\centering
%	\begin{subfigure}[b]{1\linewidth}
%		\includegraphics[width=\linewidth]{./pics/planning2}
%	\end{subfigure}
%	\caption{Visualization of the planning process for registration.}
%	\label{planning}
%\end{figure}

\section{Conclusion}
In this paper, we proposed a novel registration method by planning with a learned latent dynamic model.
In detail, we first designed an effective training framework to learn the latent dynamic model consisting of a transformation function and a reward function. Then, with the learned dynamic model, we utilized the trial-and-error based cross entropy method to search for the optimal transformation, where the dynamic model is utilized to evaluate the output of each trial and guide the evolution of CEM. Extensive experimental results on ModelNet40 and 7Scene benchmark datasets demonstrate that our method can yield good performance. %registration performance.

\section*{Acknowledgments}
This work was supported by the National Science Fund of China (Grant Nos. U1713208, 61876084, 61876083), Program for Changjiang Scholars.

\small
\bibliographystyle{named}
\bibliography{ijcai21}

\end{document}